\documentclass{article}
\usepackage[utf8]{inputenc}
\usepackage[english]{babel}
\usepackage{amsmath,amssymb,amsthm}
\usepackage{graphicx,xcolor}
\usepackage{import}
\usepackage{hyperref}
\usepackage{multicol} 
\usepackage{mathtools}
\usepackage{booktabs}
\usepackage[hypcap=false]{caption}
\usepackage{graphicx,subcaption}
\usepackage{float}
\usepackage{authblk}
\usepackage{nicefrac}
\usepackage[vlined,ruled]{algorithm2e}
\usepackage{etoolbox}
\usepackage[left=3cm, right=3cm, top=2.3cm,bottom=3.4cm]{geometry}
\usepackage{xfrac}
\usepackage{csquotes}

\usepackage[bibstyle=ieee, citestyle=numeric-comp]{biblatex}
\providecommand{\keywords}[1]
{
	\small	
	\textbf{\textit{Keywords---}} #1
}
\hypersetup{colorlinks, citecolor=red}

\newtheorem{step}{Step}
\AfterEndEnvironment{step}{\noindent\ignorespaces}

\AfterEndEnvironment{definition}{\noindent\ignorespaces}

\newcommand\pef[1]{(\ref{#1})}

\begin{document}
	\flushbottom

	\title{Solving Falkner-Skan type equations via Legendre and Chebyshev Neural Blocks}

        \author[1]{Alireza Afzal Aghaei}
        \author[1]{Kourosh Parand*}
	\author[1]{Ali Nikkhah}
	\author[1]{Shakila Jaberi}
       
	\affil[1]{Department of Computer and Data Science, Faculty of Mathematical Sciences, Shahid Beheshti University, Tehran, Iran.}

	\maketitle

	\begin{abstract}
 In this paper, a new deep-learning architecture for solving the non-linear Falkner-Skan equation is proposed. Using Legendre and Chebyshev neural blocks, this approach shows how orthogonal polynomials can be used in neural networks to increase the approximation capability of artificial neural networks. In addition, utilizing the mathematical properties of these functions, we overcome the computational complexity of the backpropagation algorithm by using the operational matrices of the derivative. The efficiency of the proposed method is carried out by simulating various configurations of the Falkner-Skan equation.
	\end{abstract} 

		
\section{Introduction }
Differential equations play an essential role in modeling natural and scientific phenomena, such as mechanics, biology, and engineering physics. Therefore, it has usually been an interesting topic for researchers. Neural networks and deep learning techniques have proven to be effective approaches for solving differential equation models \cite{hajimohammadi2020new}. Essentially, a neural network comprises multiple processing units that work in parallel and are arranged in a sequential manner. The first layer acquires the raw input data, similar to how the optic nerves function in human vision, while the final layer produces the output after processing the input data. These types of networks usually have a fixed architecture and variable weights. Scientists have successfully integrated these popular methods for solving differential equations with neural networks to achieve superior outcomes \cite{parand2011numerical, parand2012new}. 

Among the various numerical methods employed for solving functional equations, the method of weighted residuals stands out as a prominent technique within this field. This method involves approximating the solution as a linear combination of spectral basis functions, such as Jacobi polynomials, trigonometric functions, etc. Compared to finite element and finite difference methods, spectral methods possess exceptional properties of high spatial accuracy for well-behaved problems. This makes them highly suitable for numerical simulations aimed at predicting flows with a broad range of dynamically significant scales of motion. 

Jacobi polynomials and their special cases, Legendre and Chebyshev polynomials, have been used by various researchers to accelerate the numerical solution. These functions have the orthogonality property along with boundedness and symmetry which makes them powerful to handle various physical problems. In addition, the derivative of these polynomials is defined based on themselves. This property resulted in the operational matrices of derivatives\cite{sahlan2017operational, razzaghi2001legendre,parand2023neural}. These matrices can be utilized to simplify solving mathematical problems or accelerate the learning process of neural networks with orthogonal layers. In this paper, we focus on the second one and develop the Legendre and Chebyshev neural blocks (LegendreBlock and ChebyshevBlock) for the simulation of the non-linear differential equations. We will emphasize the validity of the proposed network by using several settings of the Falkner-Skan non-linear differential equation. The main contribution of our work can be summarized as follows:
\begin{itemize}
    \item Introducing the neural Legendre and Chebyshev Blocks
    \item Designing deep neural networks based on them 
    \item Using operational matrices of the derivative
    \item Solving various Falkner-Skan problems
\end{itemize}
The subsequent sections of the paper are arranged as follows: 
In Section 1, an explanation of the Blasius equation is provided, followed by the definition of the General Falkner-Skan model.
Section 2 elaborates on the formulation and operational matrix of Legendre and Chebyshev functions, which are prerequisites for the presented architecture. In Section 3, we provide a comprehensive explanation of our deep neural network architecture and discuss its application in solving the General Falkner-Skan equations using Legendre and Chebyshev blocks. The results and comparisons of our method are presented in Section 4. Finally, Section 5 is devoted to some conclusions.

\subsection{Blasius Equation}
This equation is related to the boundary layer problem, a fundamental concept in transfer phenomena. It refers to a fluid layer near a surface with high viscosity effects. The measure of a fluid's resistance to flow, known as viscosity, is an important parameter to consider for both liquids and gaseous fluids. The boundary layer is a fluid that is influenced by Heat-Motion size or mass transfer from a standard surface that can be stationary or moving. For example, we can refer to a layer of fluid that is located near a heat source, such as a radiator. The layer near the heat source creates a temperature profile by absorbing heat, which changes fluid density and initiates convection flow. Altogether, we have three types of boundary layers; Speed, mass transfer, and heat transfer. If all three phenomena coincide, then it imposes computational complexity. Nondimensional numbers can be used to express mathematical relationships between three boundary layers.
We present some mathematical models of this problem. \cite{howarth1938solution, schlichting2003boundary, dresner1983similarity, aminikhah2010exact}. The Blasius equation is given by:

\begin{equation}
\label{eq:blasius}
\frac{d^3f}{d\eta^3} + \frac{1}{2}f(\eta)\frac{df}{d\eta} = 0,
\end{equation}
where $f(\eta)$ represents the stream function, and $\eta$ is the similarity variable defined as $\eta = \frac{y}{\sqrt{x}}$, with $x$ and $y$ denoting the streamwise and wall-normal coordinates, respectively.

The boundary conditions associated with the Blasius equation are given by:

\begin{equation}
\label{eq:blasius_bc}
f(0) = f'(0) = 0, \quad f'(\infty) = 1,
\end{equation}
where $f'(x)$ denotes the derivative of $f(\eta)$ with respect to $\eta$.

    \subsection{Falkner-Skan Equation}

\begin{center} 
\begin{figure}[ht]
\centering
\caption{A boundary layer flow over a wedge}
\label{img:fig1}
\includegraphics[width=.43\textwidth]{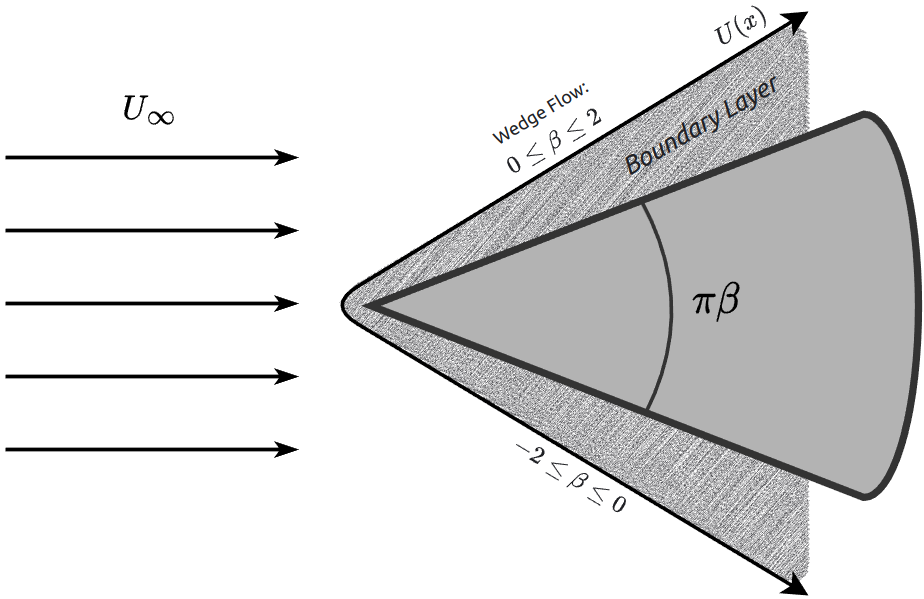}
\end{figure}
\end{center}

The Falkner-Skan equation is a non-linear boundary value problem that arises in the study of fluid mechanics. It is a second-order ordinary differential equation that describes the boundary layer flow over a semi-infinite flat plate. The equation is named after the researchers Falkner and Skan, who introduced it in the early 1940s. It has various applications in aerodynamics and heat transfer, particularly in the analysis of laminar and turbulent boundary layers. It provides valuable insights into the flow characteristics and the determination of important quantities such as the velocity and temperature profiles near the solid surface. The solution of the Falkner-Skan equation involves determining the characteristics of the laminar boundary layer on an unbounded wedge with a vertex angle of $\pi\beta$ for $ 0 \leq \beta \leq 2$, as shown in Figure \ref{img:fig1}.

Simplifying the mass continuity  and momentum equations in this case, results in the following ODE named Falkner-Skan model:

\begin{equation}
\label{eq:FS_trans_2}
    g'''(x)+\alpha g(x)g''(x)+\beta(1-(g'(x))^2) = 0,
\end{equation}
with the boundary conditions
\begin{equation}
\label{eq:FS_trans_3}
    g(0)=g'(0)=0, \quad and \quad g'(\infty)=1.
\end{equation}

\section{Orthogonal Polynomials}
Orthogonal polynomials are a unique class of polynomial functions that are characterized by a specific inner product defined on a finite or infinite interval \cite{rad2023learning}. The most widely recognized types of orthogonal polynomials comprise the Chebyshev polynomials, the Hermite polynomials, and the Legendre polynomials. This section will present a brief summary of the Legendre and Chebyshev polynomials, along with their corresponding operational derivative matrix.
    \subsection{Legendre Polynomials}
Legendre polynomials are a sequence of orthogonal functions that can be obtained by applying the Gram-Schmidt orthogonalization process to the Taylor basis functions $1, x, x^2, \dots$ \cite{rad2023learning}. These polynomials are widely employed in scientific research and have recently been employed to enhance the capability of neural networks \cite{voelker2019legendre}. Legendre polynomials can be seen as the eigenfunctions of the Sturm-Liouville problem:
  \begin{equation}
\frac{\text{d}}{\text{d}x} ((1-x^2) \frac{\text{d}}{\text{d}x} L_n(x)) + \lambda_n L_n(x) = 0,
  \end{equation}
here $\lambda_n=n(n+1)$ is the corresponding eigenvalue. 
Different intuitions can calculate these polynomials, for example, a simple one is the recurrence formula:
  \begin{equation}
    \label{eq:LEG}
\begin{aligned}
&L_{0}(x)=1, \quad L_{1}(x)=x, \\
&L_{n+1}(x)=\left(\frac{2 n+1}{n+1}\right)xL_{n}(x)-\left(\frac{n}{n+1}\right)xL_{n-1}(x), \quad n \geqslant 1 .
\end{aligned}
  \end{equation}

One key property of the Legendre polynomials is their orthogonality with respect to the $\mathcal{L}^2$-inner product, which can be expressed as:
\begin{equation}
\int_{-1}^{1}L_m(x)L_n(x)dx=\frac{2}{2n+1}\delta_{mn} ,
  \end{equation}
where $\delta_{mn}$ denotes the Kronecker delta. This orthogonality property helps us to compute its derivatives based on previous terms in the sequence. In the following, we explain this feature.
    
    \subsubsection{Operational Matrix}
The operational matrix based on the Legendre polynomial method is a recent development that has proved useful in solving mathematical problems, especially differential equations. Mathematically, this matrix is defined as:
    \begin{equation}
    \label{eq:legmatrix}
        \frac{d}{dx}L(x)=M^{(1)}L(x),
    \end{equation}
where $L(x)=[L_0(x),L_1(x),\dots,L_n(x)]^T$ is a vector containing a sequence of Legendre polynomials and $M^{(1)}$ is an $(N+1) \times(N+1)$. The entries of this matrix can be computed using the orthogonality property of these functions. This matrix can be precomputed using the following formula:

\begin{equation*}
	M^{(1)} =
	\begin{cases}
	\begin{aligned}
	&\displaystyle 2j+1, &\ i>j,\, i+j \text{ is odd} ,\\
	\displaystyle & 0, & otherwise.
	\end{aligned}
	\end{cases}
\end{equation*}		
For instance, for $N=6$, the matrix $M^{(1)}$ can be represented as:
    \begin{equation}
        M^{(1)}=
        \begin{bmatrix}
            0 & 0 & 0 & 0 & 0 & 0\\
            1 & 0 & 0 & 0 & 0 & 0\\
            0 & 3 & 0 & 0 & 0 & 0\\            
            1 & 0 & 5 & 0 & 0 & 0\\
            0 & 3 & 0 & 7 & 0 & 0\\
            1 & 0 & 5 & 0 & 9 & 0\\
        \end{bmatrix}.
    \end{equation}
The higher derivatives of $L(x)$ can be simply expressed as:
  \begin{equation}
      \frac{d^n}{dx^n}L(x)=(M^{(1)})^nL(x)=M^{(n)}L(x) \quad where \quad n=1,2,... \hspace{0.1cm} .
  \end{equation}

   \subsection{Chebyshev Polynomials}
   
Chebyshev polynomials are another family of orthogonal polynomials that were introduced by the Russian mathematician Pafnuty Chebyshev in 1854. The first kind of Chebyshev polynomials can be computed recursively using the following recurrence relation \cite{guo2002chebyshev}:

	\begin{equation}
	\label{eq:CHE}
\begin{aligned}
&C_{0}(x)=1, \quad C_{1}(x)=x, \\
&C_{n+1}(x)=2xC_{n}(x)-C_{n-1}(x), \quad n \geq 1 .
\end{aligned}
	\end{equation} 
The orthogonality property of these polynomials can be seen with respect to weight function $w(x)=1 / \sqrt{1-x^{2}}$:
  \begin{equation}
\begin{aligned}
\int_{-1}^{1}C_n(x)C_m(x)\frac{dx}{\sqrt{1-x^2}}= 	\begin{cases}
	\begin{aligned}
	 0 &\quad :n\neq m\\
	 \pi &\quad :n=m=0\\
	 \frac{\pi}{2} &\quad :n=m\neq0
	\end{aligned}
	\end{cases}
\end{aligned}
  \end{equation}
The Chebyshev polynomials also satisfy the following second-order linear Sturm-Liouville differential equation:
	\begin{equation}
\begin{aligned}
(1-x^2)\frac{\text{d}^2}{\text{d}x^2} L_n(x) - x \frac{\text{d}}{\text{d}x} L_n(x) + n^2 L_n(x) = 0,
\end{aligned}
	\end{equation} 



\subsubsection{Operational Matrix}
The operational matrix of the derivative for the Chebyshev polynomials $C(x)$ will be defined as:

\begin{equation}
    \label{eq:chmatrix}
        \frac{d}{dx}C(x)=M^{(1)}C(x).
    \end{equation}
where $C(x)=[C_0(x),C_1(x),\dots,C_n(x)]^T$ and $M^{(1)}$ is an $(N+1) \times(N+1)$ matrix defined by:
\begin{equation*}
	M^{(1)} =
	\begin{cases}
	\begin{aligned}
	\frac{2i}{c_j}, &\quad i>j,\, i+j \text{ is odd} ,\\
	0, &\quad otherwise.
	\end{aligned}
	\end{cases}
\end{equation*}		
where
\begin{equation*}
	c_k =
	\begin{cases}
	\begin{aligned}
	1, &\quad l=1,\dots,N ,\\
	2, &\quad l=0.
	\end{aligned}
	\end{cases}
\end{equation*}		 
For instance, if $N=6$, the matrix $M^{(1)}$ can be represented as:

\begin{equation}
    M^{(1)}=
    \begin{bmatrix}
    0 & 0 & 0 & 0 & 0 & 0\\
    1 & 0 & 0 & 0 & 0 & 0\\
    0 & 4 & 0 & 0 & 0 & 0\\
    3 & 0 & 6 & 0 & 0 & 0\\
    0 & 8 & 0 & 8 & 0 & 0\\
    5 & 0 & 10 & 0 & 10 & 0 
    \end{bmatrix}.
\end{equation}
Also the higher derivatives of $C(x)$ can be expressed as:
  \begin{equation}
      \frac{d^n}{dx^n}C(x)=(M^{(1)})^nC(x)=M^{(n)}C(x) \quad where \quad n=1,2,... \hspace{0.1cm} .
  \end{equation}

\section{Methodology}
\begin{figure}[ht]
	\centering
        \caption{Legendre and Chebyshev Blocks}
        \label{img:blocks}
        \begin{subfigure}{.5\textwidth}
		\centering
		\includegraphics[width=.9\linewidth]{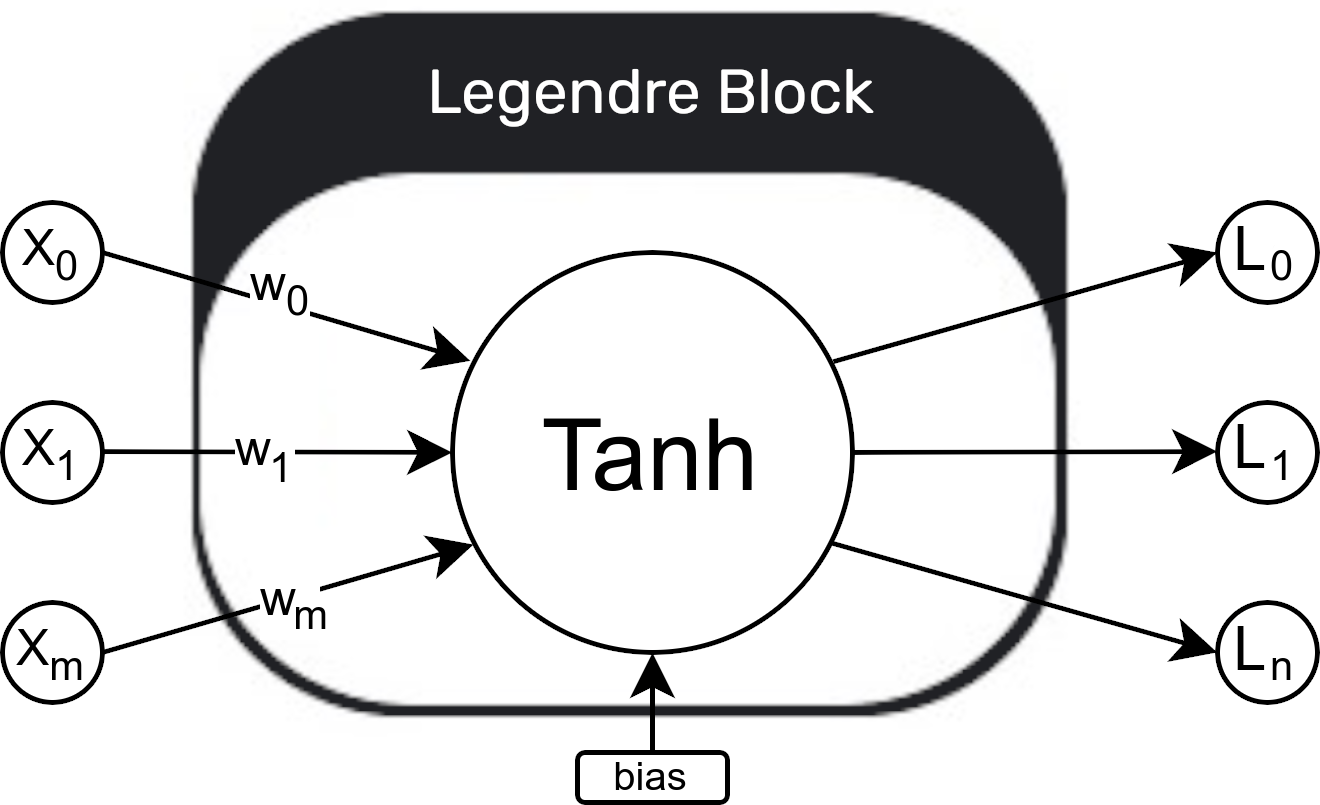}
		\caption{}
	\end{subfigure}%
	\begin{subfigure}{.5\textwidth}
		\centering
		\includegraphics[width=.9\linewidth]{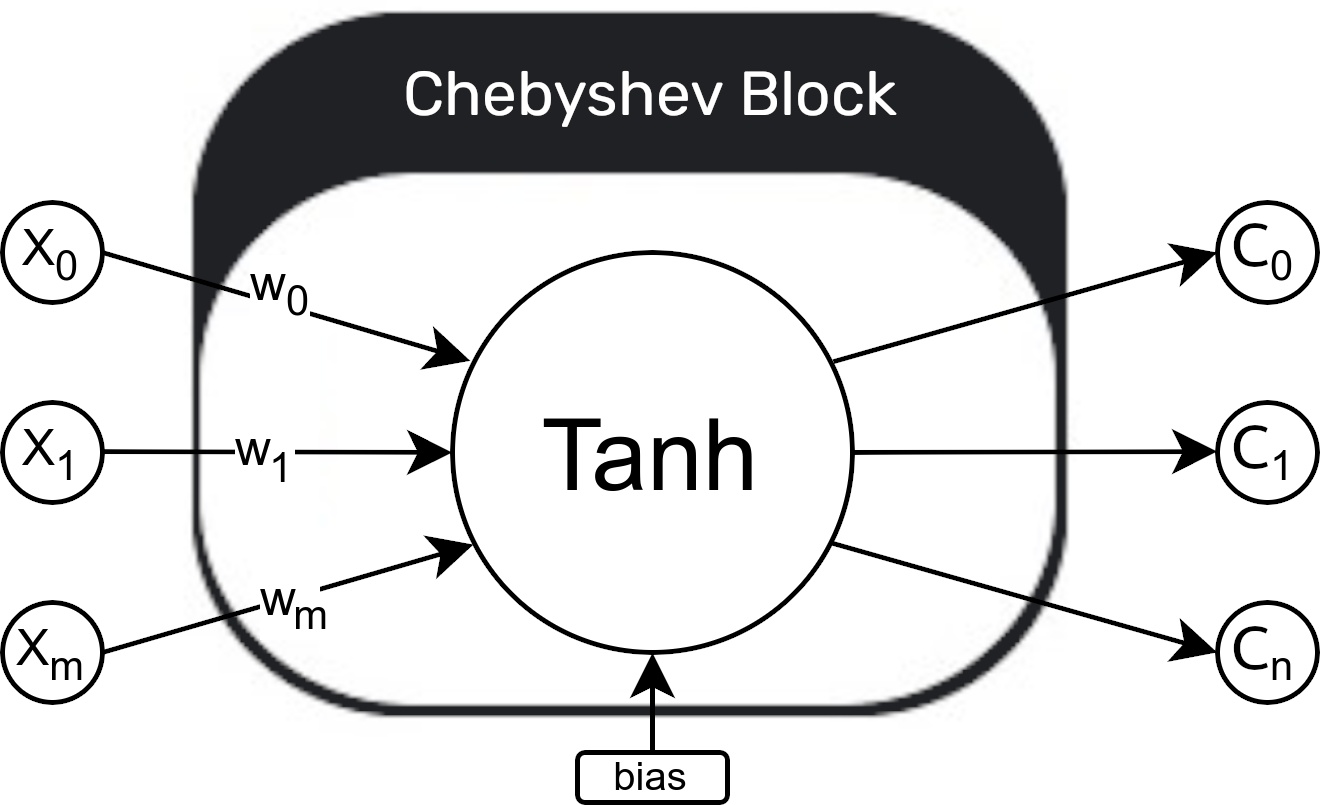}
		\caption{}
	\end{subfigure}
\end{figure}

\begin{center}
   \begin{figure}[ht]
     \centering
        \caption{Legendre-Chebyshev Deep Neural Network Architecture}
	\label{img:fig3}
     \includegraphics[width=1\textwidth]{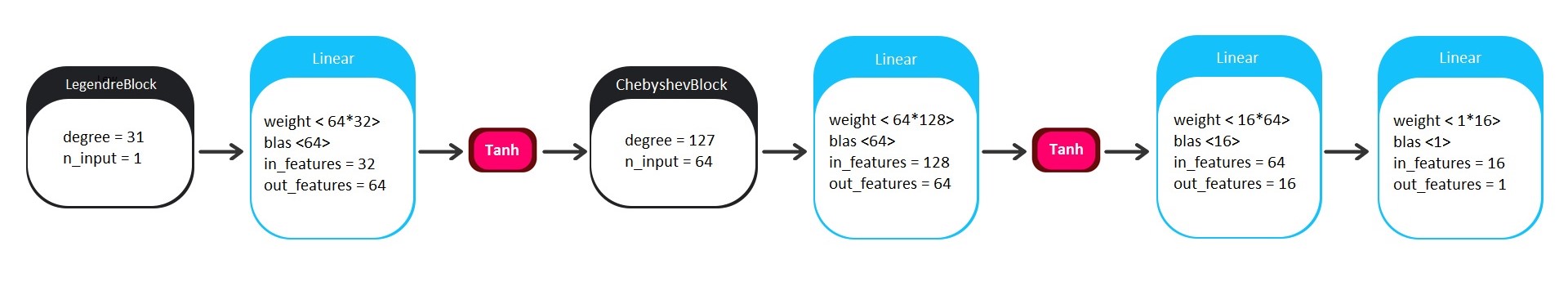}
\end{figure}
 \end{center}

The previous sections explained the prerequisites for the paper. In the first section, we introduced the Blasius and Falkner-Skan equations, then discussed the Legendre and Chebyshev polynomials. Now we explain our proposed method and finally present an architecture to solve the equation. Here, we suggest using orthogonal functions in the network. Recently orthogonal functions have been used in machine learning for approximating differential equations. Support vector machines and neural networks are some of the well-known efforts \cite{mall2016application, hadian2020single, hajimohammadi2021fractional,parand2023solving}. The applications of orthogonal functions in neural networks achieved more accurate results. In most articles, these functions have been employed as activation functions of neurons. However, they may face some difficulties. For example, the domain of these functions is bounded. This issue has been handled by using them at the first layer of the network or after the normalization technique. We present a new architecture that addresses these problems. This method presents a neural block that includes orthogonal functions and other non-linear neurons. The input and the output of the block are vectors. The input vector is encoded into a scalar using a neuron with an appropriate activation function. Then the $n-th$ order polynomials are generated by utilizing the output of this neuron. In our case, we have used Legendre and Chebyshev, whose domain is $[-1,1]$. So the activation function can be considered as the hyperbolic tangent. By doing this, we will esure that the domain of the function is satisfied. A graphical design of these blocks with Legendre and Chebyshev polynomials can be seen in figure \ref{img:blocks}. It will not have a computational overhead. That is because the forward phase evaluates the polynomials with a fixed small order. Moreover, the backward phase uses the operational matrices of the derivative instead of the backpropagation algorithm. These neural blocks can be used anywhere in a deep neural network. Figure \ref{img:fig3} shows an example network that employs these blocks in this architecture. We utilized this model in the rest of the work. This architecture can be used to learn classification or regression tasks. In addition, it can learn the hidden dynamics of a differential equation. This is done by defining the equation's residual as the network's loss function. This paper focuses on solving the Falkner-Skan type differential equation using this network. In the following, we will describe the work in detail and present the proposed Algorithm \ref{alg:ode}.

\begin{algorithm}[ht]
	    \DontPrintSemicolon
		\SetAlgoLined
		\SetKwInOut{Input}{Input}
		\SetKwInOut{Initialize}{Initialize}
		\SetKwInOut{Set}{set}
		\SetKwInOut{Compute}{compute}
		\SetKwInOut{Define}{define}
		\SetKwInOut{Solve}{sovle}
		\SetKwInOut{Output}{Output}
		\SetKwFunction{Floss}{GetLoss}
		\SetKwProg{Fn}{Function}{:}{}

		\Input{$\alpha$   and $\beta$}
		\Input{$D$ \textbf{As} domain}
		\Input{$n$ \textbf{As} number of discretization}
		\Input{Optimizers parameters $epoch_{a}, lr_{a}, epoch_{l}, lr_{l}, eps$}

        \Initialize{$X \gets GenerateTrainingSet(D,n)$ \\ $M \gets LegendreChebyshevModel()$ \tcp*[r]{Based on Figure \pef{img:fig3}}}

		\Fn{\Floss{$X$}}{
            \Compute{Derivatives $g'(x),g''(x),g'''(x)$}
            \Compute{Boundaries \tcp*[r]{Based on Eq. \pef{eq:FS_trans_3}}}
            \Compute{Residual \tcp*[r]{Based on Eq. \pef{eq:residual}}}
            \Compute{Loss \tcp*[r]{Based on Eq. \pef{eq:loss}}}
            \KwRet Loss
		}
		\vspace{1em}
		\For{$i=0$ \textbf{to} $epoch_{a}$} {
            \Set{$loss \gets GetLoss(X)$ \textbf{As} initial solution}
			\Solve{$AdamOptimizer(lr_a)$}
			\Set{obtained solutions to $X$}
		}
		\vspace{0.4em}
	    $previous \gets 0$ \\
		\For{$i=0$ \textbf{to} $epoch_{l}$} {
		    \Set{$loss \gets GetLoss(X)$}
			\Solve{$LBFGSOptimizer(lr_{l})$}
			\Set{obtained solutions to $X$}
			\If{$abs(previous - loss) < eps$}{
			\tcp{Converged}
            \textbf{break}
            }
	    $previous \gets loss$ \\
		}
		\Output{solutions $X$}
		
		\caption{General Falkner-Skan ODE algorithm for Eq. \pef{eq:FS_trans_2}. }
		\label{alg:ode}
	\end{algorithm}

\begin{step}
To fit the network, we need a training data set. This data set for a specific interval, possibly an unbounded domain, can be produced in different ways, such as equidistant points and random nodes. In spectral analysis, these are known as collocation points. 
\end{step}
\begin{step}
In our case, the dimensionality of the input and output is one. However, the proposed model can be easily adjusted for solving multi-dimensional problems such as partial differential equations. Figure \ref{img:fig3} shows the network we have used for our simulation. This model is obtained by running a hyperparameter optimization algorithm.
\end{step}
\begin{step}
Now we can define the residual function of the \eqref{eq:FS_trans_2} according to the boundary conditions \pef{eq:FS_trans_3} and the loss function: 
\begin{equation}
\begin{aligned}
\label{eq:residual}
 Residual(x) = g'''(x)+\alpha g(x)g''(x)+\beta(1-(g'(x))^2).
\end{aligned}
\end{equation}
\begin{equation}
\label{eq:loss}
 Loss(x) = \frac{1}{n}\sum_{i}{Residual(x_i)^2} + [g(0)^2+g'(0)^2+g'(\infty)^2].
\end{equation}  
\end{step}
\begin{step}
In this step, the network should be trained. These can be done using a first or second-order gradient-based optimization technique. In this paper, we recommend a two-stage approach. To do these, we first use the Adam optimizer to find suboptimal weights. In the second stage, the LBFGS algorithm with the initial weights obtained in the previous stage. 
\end{step}
\section{Results and Discussion}
This section presents the application of two proposed models, namely Legendre Deep Neural Network (LDNN) that only employs the Legendre Block, and Legendre-Chebyshev Deep Neural Network (LCDNN), to obtain solutions of the General Falkner-Skan equation for different values of $\alpha$ and $\beta$. Notably, the flows mentioned in Table \ref{tbl:flows} are widely examined and correspond to the $(\alpha, \beta)$ pairs in equation \pef{eq:FS_trans_2}.

\begin{center}
\setlength{\arrayrulewidth}{0.5mm}
\setlength{\tabcolsep}{18pt}
\renewcommand{\arraystretch}{1.4}
\captionof{table}{Some famous particular types of Falkner-Skan flows.}
\label{tbl:flows}
\begin{tabular*}{.8\linewidth}{@{\extracolsep{\fill} }lcc@{}}
			\toprule
			Type of Flow &  $\alpha$  & $\beta$ \\ 
			\midrule
                 Blasius \cite{blasius1907grenzschichten} & $\sfrac{1}{2}$  &  0 \\
                 Pohlhausen \cite{pohlhausen1921naherungsweisen} & 0 & 1 \\
                 Homann \cite{homann1936uss} & 2 & 1 \\
                 Hiemenz \cite{hiemenz1911grenzschicht} & 1 & 1 \\
                 Hastings \cite{hastings1972reversed} &  1 & $[-0.18, 2]$ \\
                 Craven \cite{craven1972reverse} & 1  &  $[10, 40]$ \\ 
            \bottomrule
\end{tabular*}
\end{center}

We consider domain $[0, 6]$ with $18000$ discrete points as the training set. We used the settings shown in Table \ref{tbl:opt} as optimizers parameters. The obtained values of $g''(0)$ for listed flows and corresponding errors obtained by the present methods and the comparison with the results of previous methods at Blasius flow where $\alpha=0.5$ and $\beta=0$ are shown in Table \ref{tbl:l1}. Also, the graphs of the loss and residual functions and the comparison between exact values with our predicted values for this case are shown in Figure \ref{fig:graphs}.

\begin{center}
\setlength{\arrayrulewidth}{0.5mm}
\setlength{\tabcolsep}{18pt}
\renewcommand{\arraystretch}{1.5}
    \captionof{table}{Optimizers parameters}
    \label{tbl:opt} 
    
    \begin{tabular*}{.6\linewidth}{@{\extracolsep{\fill} }lccc@{}}
    \toprule
    Optimizer & Epoch & Learning Rate & $eps$ \\
    \midrule
    Adam &  100 & 0.015 & 1e-10 \\
    LBFGS & 10000& 0.015035 & 1e-10 \\ 
    \bottomrule
    \end{tabular*} 
\end{center}

Furthermore, we conducted various experiments to confirm the effectiveness of LCDNN in solving the Falkner-Skan problem in the Blasius Flow. To determine the accuracy of the numerical results, we computed the error, for $n$ test data, using the following criteria:
\begin{equation*}
\begin{aligned}
\label{eq:mse}
 MSE = \frac{1}{n}\sum_{i=1}^{n} (y_i - y_i^{pred})^{2} , \\
 MAE = \frac{1}{n}\sum_{i=1}^{n} |y_i - y_i^{pred}| .
\end{aligned}
\end{equation*}
\begin{table}[htb]
\setlength{\arrayrulewidth}{0.5mm}
\setlength{\tabcolsep}{18pt}
\renewcommand{\arraystretch}{1.5}
    \caption{(a) Solutions and (b) Errors for Falkner-Skan problem at Blasius-Flow}.
    \label{tbl:l1}
    \begin{subtable}{.5\linewidth}
      \centering
        \caption{}
    \begin{tabular*}{.8\linewidth}{@{\extracolsep{\fill} }ll@{}}
			\toprule
			Method & $g^{\prime \prime}(0)$ \\ 
			\midrule
			    LCDNN & $0.3320590$ \\ 
			    LDNN & $0.3321312$ \\ 
                K. Parand \cite{parand2012new} & $0.3320573$ \\
                L. Howarth \cite{howarth1934calculation} & $0.3320600$ \\
                R. Cortell \cite{cortell2005numerical} & $0.3320600$ \\
                Boyd \cite{boyd2008blasius} & $0.3320573$ \\
            \bottomrule
	\end{tabular*}
    \end{subtable}%
    \begin{subtable}{.5\linewidth}
      \centering
        \caption{}
    \begin{tabular*}{.8\linewidth}{@{\extracolsep{\fill} }lc@{}}
    \toprule
    Error Name &        Value \\
    \midrule
        $MSE$ &  8.942052e-08 \\
        $l^1$-norm &  8.348959e-02 \\
        $l^2$-norm &  5.179397e-03 \\
        $l^\infty$-norm &  3.911157e-04 \\
        $MAE$ &  2.782986e-04 \\
    \bottomrule
    \end{tabular*}
\end{subtable} 
\end{table}
\label{fig:graphs}
\begin{figure}[ht]
	\centering
    \caption{(a) Graph of loss function vs. epoch. (b) Graph of residual vs. x. (c) Comparison of exact values and predicted values.}
	\begin{subfigure}{.5\textwidth}
		\centering
		\includegraphics[width=1\linewidth]{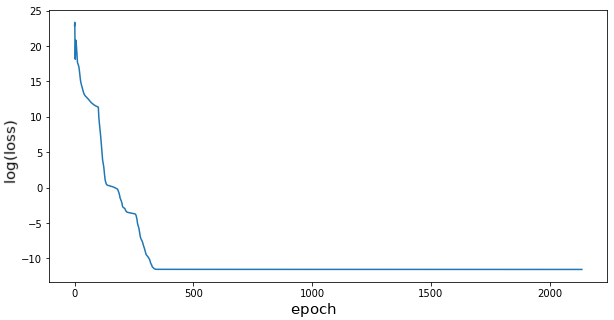}
		\caption{}
		\label{img:5}
	\end{subfigure}%
	\begin{subfigure}{.5\textwidth}
		\centering
		\includegraphics[width=1\linewidth]{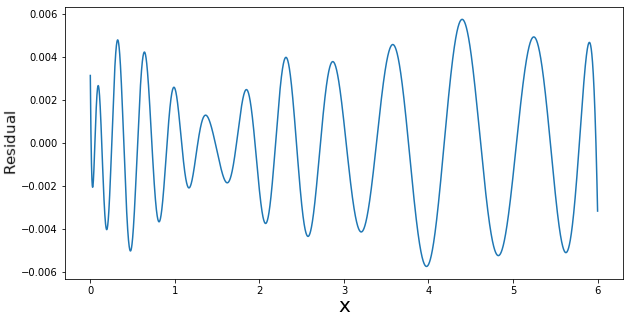}
		\caption{}
		\label{img:6}
	\end{subfigure}
	\begin{subfigure}{1\textwidth}
		\centering
		\includegraphics[width=0.8\linewidth]{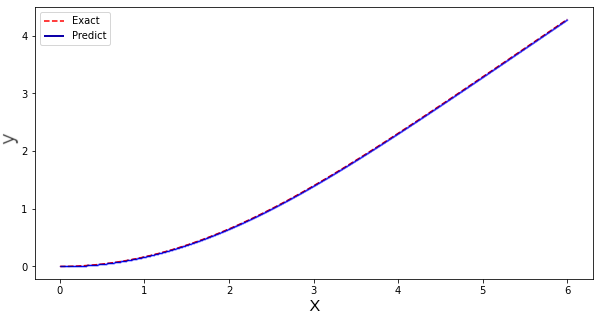}
		\caption{}
		\label{img:7}
	\end{subfigure}
\end{figure}

For Pohlhausen-Flow where $\alpha=0$ and $\beta=1$, Ames \cite{ames1968nonlinear} shows the solution as
\begin{equation*}
\begin{aligned}
\label{eq:pohlsol}
 g''(0) = \frac{2}{\sqrt{3}} \approx 1.154700538 .
\end{aligned}
\end{equation*}

Table \ref{tbl:bench} compares the benchmark solutions for various flows listed in Table \ref{tbl:flows} using both the LDNN and LCDNN methods. The results demonstrate that the LCDNN method exhibits greater accuracy than the LDNN method, and the solutions obtained by LCDNN are consistent with the previously computed solutions of the Falkner-Skan problem.\\

Tables \ref{tbl:l2} and \ref{tbl:l3} present the obtained values of $g''(0)$ using our proposed methods along with a comparison with the QLM method in \cite{hajimohammadi2020new}, Taylor method in \cite{asaithambi2005solution}, and Higher-order method in \cite{salama2004higher}. The results are presented for various values of $\alpha$ and $\beta$.

\begin{center}
\setlength{\arrayrulewidth}{0.5mm}
\setlength{\tabcolsep}{18pt}
\renewcommand{\arraystretch}{1.5}

\captionof{table}{Comparison of $g^{\prime \prime}(0)$ values obtained by the present methods with standard solutions for each flow.}
\label{tbl:bench}

\begin{tabular*}{1\linewidth}{@{\extracolsep{\fill} }lccc@{}}
			\toprule
			Type of Flow  & LCDNN  & LDNN & $g''(0)$  \\ 
			\midrule
                 Pohlhausen \cite{pohlhausen1921naherungsweisen} & 1.154709 & 1.154779 & 1.154700 \\
                 Homann \cite{homann1936uss} & 1.311937 & 1.312304 & 1.311938 \\
                 Hiemenz \cite{hiemenz1911grenzschicht} & 1.232548 & 1.230178 & 1.232589 \\
            \bottomrule
\end{tabular*}

\captionof{table}{Comparison of $g^{\prime \prime}(0)$ values obtained by the present methods with previous results where $\alpha=1$ and $\beta \in [-0.18,2]$.}
\label{tbl:l2}

\begin{tabular*}{1\linewidth}{@{\extracolsep{\fill} }ccccc@{}}
			\toprule
			$\beta$ &       LCDNN  & LDNN & Taylor\cite{asaithambi2005solution} & Higher-order\cite{salama2004higher} \\ 
			\midrule
                -0.1800 &  0.128635 &  0.125828 & 0.128637 & 0.128638 \\
                -0.1000  &  0.319270 &  0.319447 & 0.319270 & 0.319270 \\
                 0.0000 &  0.469600 &  0.469872 & 0.469600 & 0.469600 \\
                 0.5000 &  0.927680 &  0.927891 & 0.927680 & 0.927680 \\
                 1.0000 &  1.232590 &  1.232883 & 1.232589 & 1.232588 \\
                 2.0000 &  1.687218 &  1.687618 & 1.687218 & 1.687218 \\ 
            \bottomrule
\end{tabular*}

\captionof{table}{Comparison of $g^{\prime \prime}(0)$ values obtained by the present methods with previous results where $\alpha=1$ and $\beta \in [10,40]$.}
\label{tbl:l3}

\begin{tabular*}{1\linewidth}{@{\extracolsep{\fill} }cccccc@{}}
			\toprule
			$\beta$ &       LCDNN & LDNN & QLM\cite{hajimohammadi2020new} & Taylor\cite{asaithambi2005solution} & Higher-order\cite{salama2004higher} \\ 
			\midrule
                10 &  3.675234 &  3.676033 & 3.675234 & 3.675234 & 3.675234 \\
                15 &  4.491488 &  4.492465 & 4.491487 & 4.491487 & 4.491487 \\
                20 &  5.180718 &  5.181792 & 5.180718 & 5.180718 & 5.180718 \\
                30 &  6.338208 &  6.339053 & 6.338209 & 6.338209 & 6.338208 \\
                40 &  7.314785 &  7.315768 & 7.314785 & 7.314785 & 7.314785 \\
            \bottomrule
	\end{tabular*}
\end{center}

\section{Conclusion}

In this study, we have proposed an orthogonal neural block for efficiently learning the solution to non-linear differential equations. The presented architecture leverages the Chebyshev and the Legendre orthogonal polynomials as an activation function of the neurons. These polynomials benefit from various mathematical properties, including operational matrices of the derivative. This property reduces the time complexity of the backpropagation by prescribing the backward derivative operations. 

The proposed block was applied in a deep-learning architecture to solve some non-linear boundary value problems of the Falkner-Skan type. To validate the effectiveness and accuracy of this approach, various settings of this equation, including Blasius, Hastings, and Craven flow were simulated. Then several comparisons have been made with other numerical and analytical methods, and the results showed a demonstrating agreement between our results with previous works. 

The developed method was implemented using the PyTorch framework on Python $3.10$ with Graphic Processing Unit (GPU) to accelerate the learning process. As a recommended avenue for future work, we suggest developing the method to tune the network's hyperparameters and layers to further improve the learning speed and achieve even more accurate results. In addition, we suggest validating this technique to high dimensional differential equations with initial and boundary conditions.

\section{Acknowledgements}
The authors affirm that there are no competing interests to disclose in relation to the publication of this paper.

\printbibliography

\end{document}